\documentclass[twocolumn]{article}
\usepackage{M-PSI} %
\removefirstpagelogo{}

\usepackage{lipsum} 
\usepackage{multirow}
\usepackage{rotating}
\usepackage{booktabs}
\usepackage{graphicx} 
\usepackage{caption}  
\usepackage{footnote}

\usepackage[pdfencoding=auto, pdfusetitle]{hyperref}
\hypersetup{
  hidelinks,
  urlcolor=red,
  pdftitle={Exploring VQ-VAE with Prosody Parameters for Speaker Anonymization},
  pdfauthor={Sotheara Leang, Anderson Augusma, Eric Castelli, Fr\'ed\'erique Letu\'e, Sethserey Sam, Dominique Vaufreydaz},
  pdfkeywords={speech anonymization, speech synthesis, vector-quantized variation auto-encoder, emotional state}
}

\title{Exploring VQ-VAE with Prosody Parameters for Speaker Anonymization}
\author{Sotheara Leang\textsuperscript{\hspace{0.2mm}1,2}, \href{https://augusmaa.github.io/}{Anderson Augusma\textsuperscript{\hspace{0.2mm}$1,3$,~\small{\ExternalLink}}}, \href{https://www.m-psi.fr/wp-content/uploads/2021/02/CV-CASTELLI-Eric.htm}{\'Eric Castelli\textsuperscript{\hspace{0.2mm}1,~\small{\ExternalLink}}}, \href{https://membres-ljk.imag.fr/Frederique.Letue/}{Fr\'ed\'erique Letu\'e$^{\hspace{0.2mm}3,~\small{\ExternalLink}}$},\And Sethserey Sam\hspace{0.2mm}$^{2}$, \href{https://research.vaufreydaz.org/}{Dominique Vaufreydaz\textsuperscript{\hspace{0.2mm}$1$,~\small{\ExternalLink}}} \vspace{0.25cm}\\
    $^1$~Univ. Grenoble Alpes, CNRS, Grenoble INP, LIG, 38000 Grenoble, France\\
    $^2$~Institute of Digital Research and Innovation, CADT, Phnom Penh, Cambodia\\
    $^3$~Univ. Grenoble Alpes, CNRS, LJK, 38000 Grenoble, France
}

\begin{document}

\begin{abstract}
Human speech conveys prosody, linguistic content, and speaker identity.
This article investigates a novel speaker anonymization approach using an end-to-end network based on a Vector-Quantized Variational Auto-Encoder (VQ-VAE) to deal with these speech components.
This approach is designed to disentangle these components to specifically target and modify the speaker identity while preserving the linguistic and emotionalcontent.
To do so, three separate branches compute embeddings for content, prosody, and speaker identity respectively.
During synthesis, taking these embeddings, the decoder of the proposed architecture is conditioned on both speaker and prosody information, allowing for capturing more nuanced emotional states and precise adjustments to speaker identification.
Findings indicate that this method outperforms most baseline techniques in preserving emotional information. However, it exhibits more limited performance on other voice privacy tasks, emphasizing the need for further improvements. 
\end{abstract}

\section{Introduction}

Preserving privacy has become a key concern in artificial intelligence research, especially due to the widespread use of deep learning architectures that rely heavily on large datasets, often containing personal information. In contemporary applications, audio speech serves various purposes, including voice recognition systems for security and accessibility, virtual assistants for personalized user interactions, customer service automation for efficient query handling, and emotion analysis for enhancing user experience. Each of these applications underscores the need for robust privacy-preserving techniques to protect individuals' sensitive information while leveraging the power of AI.
However, the inherent risk of speaker identification presents a significant threat to personal privacy. In response to this problem, the Voice Privacy Challenge 2024~\cite{tomashenko2024voiceprivacy} aims to tackle the critical task of anonymizing speech while preserving pertinent information, notably the emotional state of the speaker. Drawing on insights from previous challenges~\cite{tomashenko2024voiceprivacy, fang2019speaker, patino2020speaker, meyer2023prosody, panariello2024speaker, champion2023anonymizing},  This article depicts a novel approach using discrete representation by a vector-quantized neural network for the speaker anonymization. The approach builds upon the foundations laid by existing research, leveraging advancements in vector quantization and neural network techniques to achieve effective speaker anonymization while preserving the emotional nuances conveyed in the speech. Organizers proposed several baselines for the Voice Privacy Challenge. The proposed architecture employs a similar approach as baselines B1, B5, and B6~\cite{tomashenko2024voiceprivacy}.
In the proposed approach, the fundamental frequency (F0) and x-vectors are extracted for the purpose of anonymization, similar to the methods used in B1. Specifically, we focus on modifying the F0 component and integrating Vector-Quantized Variational Auto-Encoder (VQ-VAE). This strategy aligns with the techniques employed in B5 and B6~\cite{tomashenko2024voiceprivacy}, aiming to leverage VQ-VAE's capabilities to enhance the effectiveness of the anonymization process while preserving essential speech characteristics. The proposed architecture aims to strike a balance between two essential objectives: anonymizing speaker information to protect privacy and retaining the emotional context of speech. By employing vector-quantized neural networks, we introduce a robust framework capable of achieving both objectives simultaneously. Throughout this paper, comprehensive explanations and analysis of the approach are provided, showcasing its efficacy in anonymizing speaker identity while faithfully preserving the emotional content of the speech data.

In the subsequent sections, the technical details of the proposed architecture are delved into, with its design principles, implementation strategies, and experimental results being elucidated. The effectiveness of the approach in achieving the goals set by the Voice Privacy Challenge 2024 is demonstrated through rigorous evaluation and comparison with existing methods.

\section{Related Work}

Prior research has explored various approaches to speaker anonymization and speech synthesis, employing a range of methodologies including deep learning and statistical techniques. Many studies have utilized prosodic features such as fundamental frequency (F0) and energy, as well as speaker embedding like x-vector proposed by Snyder et al.~\cite{snyder2018x} for speaker information manipulation.

Wawalim et al.~\cite{mawalim2020x} developed an anonymization system based on F0 analysis and modified x-vectors, using Singular Value Modification and statistical regression models to enhance speaker privacy and alter speaker identifiable characteristics. Their approach underscores the importance of sophisticated statistical methods in handling speaker-related data transformations for privacy-sensitive applications. Gaznepoglu et al.~\cite{gaznepoglu2022voiceprivacy} investigate F0 trajectory correction with a DNN where F0 trajectory is predicted in a logarithmic scale with a global mean-variance normalization.  Champion et al.~\cite{champion2021study} apply a modification of F0 using a linear transformation based on the mean and standard deviation of log-scaled F0. In their work, the linear transformation is performed only on voiced frames. 
Meyer et al.~\cite{meyer2023prosody} explored using random offsets for prosody cloning to maintain the naturalness and variability of speech characteristics essential for preserving speaker identity nuances. Their approach complements existing methods by focusing on the nuanced manipulation of prosodic elements to ensure that synthesized speech retains authenticity and remains intelligible.

This research investigates the use a vector-quantized variational auto-encoder (VQ-VAE) combined with prosody information to improve the disentanglement of content and speaker information~\cite{chorowski2019unsupervised, van2017neural, champion2022disentangled} during speaker anonymization. Integrating prosody parameters enhances emotional expression and improves the fidelity of synthesized speech, aligning with recent advancements in speech synthesis~\cite{kong2020hifi}.

\section{Proposed Method}

\begin{figure*}[t]
  \centering
  \includegraphics[width=0.8\linewidth]{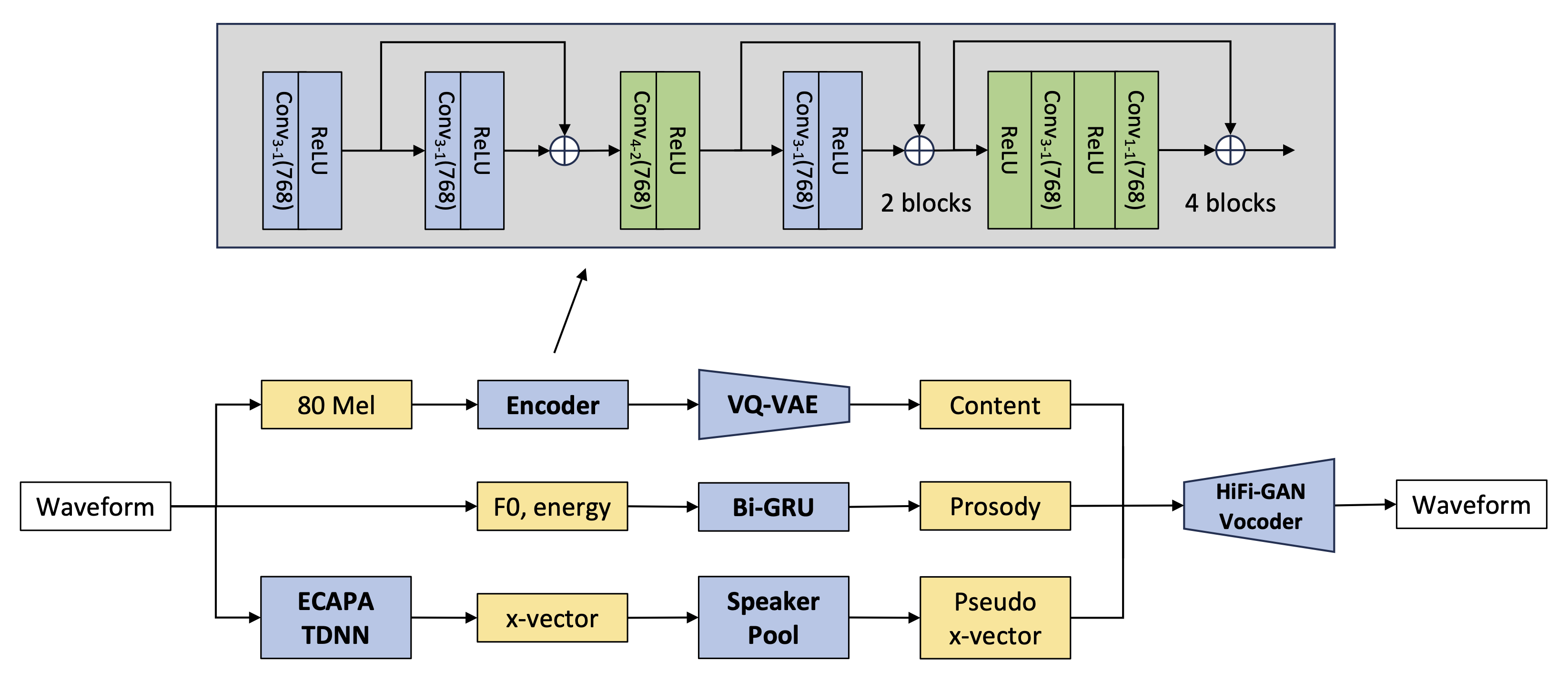}
\caption{The proposed architecture: The top figure shows the encoder of the content module, while the bottom figure depicts the anonymization system, including the content, prosody, anonymization, and decoder modules. The system takes as input 80 mel-spectrogram, F0, energy, and x-vector. The pseudo-x-vector, with content and prosody embedding, is fed to the network to produce anonymized speech.}
  \label{fig:net_all}
\end{figure*}

The proposed network employs a vector-quantized variational auto-encoder to separate speaker and content information. In addition to speaker information (x-vector), the decoder is conditioned on prosody information learned from the fundamental frequency (F0) and the energy of the spectrum. This conditioning enhances the model to focus more on the content of the speech. The detailed architecture of the proposed model is depicted in Figure~\ref{fig:net_all}.

\subsection{Content Module}

The content module comprises an encoder followed by vector quantization. The encoder includes two front-end convolution blocks, each with a kernel size of 3, a stride of 1, and 768 channels. This is followed by a downsampling convolution block with a kernel size of 4 and a stride of 2, which reduces the temporal resolution of the input feature from 100Hz to 50Hz. The sequence continues with two additional residual convolution blocks that mirror the configuration of the front-end blocks and concludes with four residual blocks, as shown in Figure~\ref{fig:net_all}. This network is based on the encoder of~\cite{chorowski2019unsupervised} except for the last four residual blocks, composed of ReLU activation functions and the convolutions. The encoder processes 80 mel-spectrograms as input and produces a 256-dimensional output representation. This output is then projected into discrete codes using a vector quantization module, which employs a codebook with 1024 codes, each 256-dimensional.

\subsection{Prosody Module}
To enhance the ability to capture the subtle nuances of intonation and emotional expression in speech, we propose incorporating two pivotal parameters: the fundamental frequency (F0), and the energy of the spectrum. These parameters are essential in enriching the prosody information supplied to the decoder, significantly improving the accuracy of speech reconstruction and elevating the efficacy of emotion detection in our model. The fundamental frequency (F0) was extracted from the audio waveform using pYAAPT\footnote{\hspace{0.5mm}\url{http://bjbschmitt.github.io/AMFM_decompy/pYAAPT.html} (last seen 08/2024).}, following the challenge guidelines. The F0 was normalized using a logarithmic scale, while the energy was normalized using the mean. The F0 and energy are then fed into a Bi-directional Gated Recurrent Unit (Bi-GRU) network with a hidden state dimensionality of 128, enabling robust temporal analysis of the prosody information.

\subsection{Anonymization Module}

The speaker anonymization process closely follows that of Baseline 1. However, the pre-trained ECAPA-TDNN~\cite{desplanques2020ecapa} was used to compute the x-vector, known for its effectiveness in capturing robust speaker characteristics. The original x-vector was replaced with a pseudo-x-vector, obtained by averaging 100 x-vectors randomly selected from the 200 most distant x-vectors based on Euclidean distance. These distant x-vectors were chosen from the speaker pool, created with the mean x-vector of the 1417 training speakers, ensuring a diverse representation of speaker traits. 

The effects of the fundamental frequency (F0) were investigated during the anonymization. Modifying the F0 can prevent the disclosure of identifiable speaker information. Firstly, we propose randomly adjusting the utterance F0 uniformly and independently between 0.8 and 1.2 to shift individual prosodic patterns while preserving the overall prosody of the utterance. Secondly, we propose normalizing the F0 using the mean of the most 100 dissimilar speakers obtained when computing the pseudo-x-vector.

\subsection{Decoder Module}

The HiFiGAN vocoder~\cite{kong2020hifi} is used as the decoder to synthesize the speech. The embedding from the content module was upsampled to 100Hz, concatenated with the embedding from the prosody module, and fed into the decoder along with the pseudo-x-vector. In this setup, the prosody embedding provides nuanced, time-varying information as local conditioning, while the pseudo-x-vector offers overarching speaker characteristics as global conditioning. The embeddings were upsampled with factors of 10, 4, and 4, totaling 160, to effectively reconstruct the waveform, guaranteeing that the synthesized speech corresponds to the original sample rate. Each upsampling stage employs kernels sized 20, 8, and 8, respectively.

\section{Experiments}

This work investigates the vector quantized-variational auto-encoder (VQ-VAE) alongside fundamental frequency (F0) in three scenarios. System 1 employs the standard normalized F0 using a logarithmic transformation. System 2 applies a random scaling factor between 0.8 and 1.2 to F0. System 3 normalizes F0 using the mean value of the speakers with the most variation within the speaker pool. The distant speakers were obtained through the same process during the computation of the pseudo-x-vector. It is important to note that all experiments and evaluations follow the challenge guidelines~\cite{tomashenko2024voiceprivacy}. The proposed system was evaluated against six baselines (B1-6) as specified in the challenge.

\subsection{Datasets}

All datasets used in the experiments complied with the challenge guidelines. The training data includes subsets from LibriSpeech~\cite{panayotov2015librispeech} and additional data from CREMA-D~\cite{cao2014crema} aiming to enhance the ability of the model to recognize and synthesize emotions. The speakers from the LibriSpeech were utilized to create a speaker pool for anonymization. Table \ref{tb_train_data} provides detailed statistical information on the composition and distribution of the training data. The development and test sets included subsets of both LibriSpeech and IEMOCAP~\cite{busso2008iemocap}. These datasets were explicitly chosen to evaluate the performance across multiple tasks, including Automatic Speaker Verification (ASV), Automatic Speech Recognition (ASR), and Speaker Emotion Recognition (SER).

\begin{table}[ht]
  \caption{Statistical information about training data.}
  \label{tb_train_data}
  \centering
  \begin{tabular}{llrr}
    \hline
    \textbf{Corpus}
        & \textbf{Subset}
        & \textbf{Hour}
        & \textbf{Speaker} \\
    \hline
    \multirow{2}{*}{LibriSpeech} & train-clean-100 & 100.6 & 251 \\
                              & train-other-500 & 496.7 & 1,166 \\
    CREMA-D &  & 5.2 & 91 \\
    \hline
  \end{tabular}
\end{table}

\subsection{Evaluation Metrics}

The anonymization systems were evaluated using three objective metrics. The Equal Error Rate (EER) served as the privacy metric, while the Word Error Rate (WER) and Unweighted Average Recall (UAR) were used as utility metrics for ASR and SER, respectively. The EER and WER were employed to assess the systems on Librispeech, and the UAR was used for the IEMOCAP test sets.

\subsection{Experimental Setup}

Two types of discriminators were used during the training process: the Multi-Period Discriminator (MPD) and the Multi-Scale Discriminator (MSD). These discriminators deal with multi-resolution concerning the temporal and local features, allowing the model to capture fine-grained speech details. The MPD and MSD follow the implementation proposed in~\cite{kim2021conditional}. The MPD was specifically simplified by targeting periods with factors of 3, 5, and 7. This modification was aimed at reducing the complexity of the discriminator, while ensuring that the model remains robust yet computationally feasible, aligning with the objectives of producing realistic and natural-sounding synthetic speech.

All the input features, including the 80 mel-spectrogram, fundamental frequency (F0), and energy, were computed using a window length of 25 milliseconds and a hop length of 10 milliseconds. The FFT size of 1024 was used to generate the spectrogram. The training was conducted over 150 epochs with a batch size of 128. The AdamW optimizer was utilized with $\beta_{1}=0.8$, $\beta_{2}=0.99$. The learning rate started at an initial value of $2 \times 10^{-4}$ and gradually decreased by 0.999 factor following each epoch. This configuration is consistent with the approach used in HiFiGAN~\cite{kong2020hifi}.

\section{Results and Discussions}

The performance of the three systems, as detailed in Tables~\ref{tab:eer_dev}, \ref{tab:eer_test}, \ref{tab:wer}, and \ref{tab:uar}, indicates lower EERs than most baselines. This suggests a somewhat reduced effectiveness in terms of privacy, which may be due to the traditional approach used to compute the pseudo-x-vector and the disentanglement challenges associated with vector quantization when the codebook size is relatively large. However, system~2 outperforms the original configuration and baselines 1 and 2. Furthermore, it achieved a better UAR in speaker emotion recognition across test sets, ranking second after baseline B2. This highlights that the method is able to retain substantial information related to emotional information.

Additionally, system~2 yielded superior results in ASR than baselines B2 and B6. The performance enhancements were particularly notable for system 1, demonstrating a lower WER than B4. This suggests that although scaling the F0 with random factor between 0.8 and 1.2 might lead to some content information loss, the overall emotional expression is primarily maintained. This indicates robustness in capturing emotional nuances.

Despite its lower performance in SER compared to systems 1 and 2, system 3 outperformed some baselines. Nevertheless, it reported the highest WER across all test sets, indicating that normalizing F0 based on the mean values of the most distant speakers adversely impacts crucial content information within the speech. This normalization process distorts essential speech characteristics, compromising speech clarity and intelligibility.

\begin{table}[ht]
    {
    \setlength{\tabcolsep}{10pt}
    \centering
    \captionsetup{singlelinecheck=off} 
    \caption{The Equal Error Rates (EER, \%) on LibriSpeech-dev achieved for male (M) and female (F) by the baselines (B1-6) and original (Orig.) data vs. the proposed systems}
    \label{tab:eer_dev}
    \begin{tabular}{lccc} 
        \toprule
        \textbf{Models} & \textbf{F $\uparrow$} & \textbf{M$\uparrow$} & \textbf{Avg$\uparrow$} \\
        \midrule
        Orig. & 10.51\hphantom{ (0)} & \hphantom{0}0.93\hphantom{ (0)} & \hphantom{0}5.72\hphantom{ (0)} \\
        
        B1 & 10.94 (9) & \hphantom{0}7.45 (5) & \hphantom{0}9.20 (7) \\
        B2 & 12.91 (8) & \hphantom{0}2.05 (9) & \hphantom{0}7.48 (9) \\
        B3 & 28.43 (3) & 22.04 (3) & 25.24 (3) \\
        B4 & 34.37 (2) & 31.06 (2) & 32.71 (2) \\
        
        \textbf{B5} & \textbf{35.82 (1)} & \textbf{32.92 (1)} & \textbf{34.37 (1)} \\
        
        B6 & 25.14 (4) & 20.96 (4) & 23.05 (4) \\
        \midrule
        System 1 & 16.47 (6) & \hphantom{0}2.79 (7) & \hphantom{0}9.63 (6) \\
        \underline{System 2} & \underline{17.91 (5)} & \hphantom{0}2.32 (8) & \underline{10.11 (5)} \\
        System 3 & 14.05 (7) & \hphantom{0}\underline{3.09 (6)} & \hphantom{0}8.57 (8) \\
        \bottomrule
    \end{tabular}
    
    } 
    
\end{table}
\footnotetext{~In all tables, for each metric, the best system is in bold, the best proposal is underlined and rank of all systems are provided.}

\begin{table}[ht]
    \centering
    {
    \setlength{\tabcolsep}{10pt}
    \caption{The Equal Error Rates (EER, \%) on LibriSpeech-test achieved for male (M) and female (F) by the baselines (B1-6) and original (Orig.) data vs. the proposed systems\textsuperscript{\hspace{0.2mm}2}.}
    \label{tab:eer_test}
    \begin{tabular}{lccc} 
        \toprule
        \textbf{Models} & \textbf{F $\uparrow$} & \textbf{M$\uparrow$} & \textbf{Avg$\uparrow$} \\
        \midrule
        Orig. & \hphantom{0}8.76\hphantom{ (0)} & \hphantom{0}0.42\hphantom{ (0)} & \hphantom{0}4.59\hphantom{ (0)} \\
        
        B1 & \hphantom{0}7.47 (8) & \hphantom{0}4.68 (5) & \hphantom{0}6.07 (6) \\
        B2 & \hphantom{0}7.48 (7) & \hphantom{0}1.56 (8) & \hphantom{0}4.52 (8) \\
        B3 & 27.92 (3) & 26.72 (3) & 27.32 (3) \\
        B4 & 29.37 (2) & 31.16 (2) & 30.26 (2) \\
        
        \textbf{B5} & \textbf{33.95 (1)} & \textbf{34.73 (1)} & \textbf{34.34 (1)} \\
        
        B6 & 21.15 (4) & 21.14 (4) & 21.14 (4) \\
        \midrule
        System 1 & \hphantom{0}8.76 (6) & \hphantom{0}\underline{2.67 (6)} & \hphantom{0}5.72 (7) \\
        \underline{System 2} & \underline{11.31 (5)} & \hphantom{0}\underline{2.67 (6)} & \hphantom{0}\underline{6.99 (5)} \\
        System 3 & \hphantom{0}6.38 (9) & \hphantom{0}2.00 (7) & \hphantom{0}4.19 (9) \\
        \bottomrule
    \end{tabular}
    } %
\end{table}

\begin{table}[ht]
    \centering
    \caption{Word Error Rates (WER, \%) achieved by the baselines (B1-6) and original (Orig.) data vs. the proposed systems\textsuperscript{\hspace{0.2mm}2}.}
    \label{tab:wer}
    \begin{tabular}{lcc}
        \toprule
        \textbf{Models} & \textbf{LibriSpeech-dev $\downarrow$} & \textbf{LibriSpeech-test $\downarrow$} \\
        \midrule
        Orig. & \hphantom{0}1.80\hphantom{ (0)} & \hphantom{0}1.85\hphantom{ (0)}\\
        \textbf{B1} & \hphantom{0}\textbf{3.07 (1)} & \hphantom{0}\textbf{2.91 (1)} \\
        B2 & 10.44 (8) & \hphantom{0}9.95 (8)\\
        B3 & \hphantom{0}4.29 (2) & \hphantom{0}4.35 (2)\\
        B4 & \hphantom{0}6.15 (5) & \hphantom{0}5.90 (6)\\
        B5 & \hphantom{0}4.73 (3) & \hphantom{0}4.37 (3)\\
        B6 & \hphantom{0}9.69 (7) & \hphantom{0}9.09 (7)\\
        \midrule
        \underline{System 1} & \hphantom{0}\underline{6.13 (4)} & \hphantom{0}\underline{5.27 (4)}\\
        System 2 & \hphantom{0}6.59 (6) & \hphantom{0}5.39 (5)\\
        System 3 & 13.65 (9) & 11.04 (9)\\
        \bottomrule
    \end{tabular}
 \end{table}

\begin{table}[ht]
    \centering
    \caption{Unweighted Average Recall (UAR, \%) achieved by the baselines (B1-6) and original (Orig.) data vs. the proposed systems\textsuperscript{\hspace{0.2mm}2}.}
    \label{tab:uar}
    \begin{tabular}{lcc}
        \toprule
        \textbf{Models} & \textbf{IEMOCAP-dev $\uparrow$} & \textbf{IEMOCAP-test $\uparrow$} \\
        \midrule
        Orig. & 69.08\hphantom{ (0)} & 71.06\hphantom{ (0)} \\
        B1 & 42.71 (4) & 42.78 (4)\\
        \textbf{B2} & \textbf{55.61 (1)} & \textbf{53.49 (1)} \\
        B3 & 38.09 (7) & 37.57 (7) \\
        B4 & 41.97 (6) & 42.78 (4) \\
        B5 & 38.08 (8) & 38.17 (5) \\
        B6 & 36.39 (9) & 36.13 (8) \\
        \midrule
        System 1 & 45.45 (3) & 44.23 (3) \\
        \underline{System 2} & \underline{45.56 (2)} & \underline{44.85 (2)} \\
        System 3 & 42.28 (5) & 38.06 (6) \\
        \bottomrule
    \end{tabular}
\end{table}

\section{Conclusion}

This work examined a vector-quantized variational auto-encoder with prosody parameters, including the fundamental frequency (F0) and the spectrum's energy for speaker anonymization. The findings demonstrate that employing a vector quantization on the variational auto-encoder to disentangle content and speaker identity involves loss of information. However, that loss does not significantly reduce the efficacy of voice conversion processes. Moreover, the introduced method outperforms most of the baselines in terms of emotion recognition. This underlines the advantages of integrating discrete representations with the prosody parameters. However, the systems perform less significantly than most baselines for speaker anonymization. 

Further investigation into the trade-offs associated with different codebook sizes could enhance disentanglement and improve anonymization. Reducing the quantizer's focus on speaker information would also be advantageous. Implementing a speaker classifier with gradient reversal as an auxiliary network on the content encoder's output could effectively penalize speaker-related content. Lastly, exploring advanced methods for generating pseudo-x-vectors could further refine anonymization. These combined efforts will contribute to more effective and nuanced anonymization strategies.

\section*{Acknowledgments}
This research was partially supported by the PERSYVAL Labex (ANR-11-LABX-0025), by the TALISMAN project (ANR-22-CE38-0007), by a French Government Scholarship (BGF), and was granted access to the HPC resources of IDRIS under the allocation 2023-AD010614233
made by GENCI. This research was made possible by the collaboration between \href{https://www.m-psi.fr}{M-PSI team\textsuperscript{\small{~\ExternalLink}}} at Grenoble Informatics Laboratory (LIG) and Cambodia Academy of Digital Technology (CADT).

\bibliographystyle{plain}
\bibliography{biblio}

\end{document}